%% file: AnomalyTracking.tex
\documentclass[conference,10pt]{IEEEtran}
\IEEEoverridecommandlockouts

\usepackage{amsmath,amsthm, amssymb,amsfonts,bbm}
\usepackage{cite,cleveref}
\usepackage{algorithm,algorithmic}
\usepackage{enumitem,xcolor}

\usepackage{graphicx}
\usepackage{caption} 
\usepackage{subcaption}	
\graphicspath{{Figures/}}
\usepackage{array}

\newcommand{\upi}{\pi_{\mathrm{upper}}}
\newcommand{\uN}{N_{\mathrm{thres}}}
\newcommand{\tchange}{t_{\mathrm{change}}}
\newcommand{\tstop}{t_{\mathrm{stop}}}
\newcommand{\bin}[2]{\calB_{#1}\lc#2\rc}

\usepackage{etoolbox}
\newcommand{\change}[1]{{#1}}
\input{My_notations}

\begin{document}
\title{Temporal Detection of Anomalies via Actor-Critic Based Controlled Sensing}

\author{%
  \IEEEauthorblockN{Geethu~Joseph, M.~Cenk~Gursoy, and Pramod~K.~Varshney~\IEEEmembership{Life~Fellow,~IEEE}}
  \IEEEauthorblockA{Department of Electrical Engineering and Computer Science\\Syracuse University, Syracuse, NY 13244, USA\\ e-mails: \{gjoseph,mcgursoy,varshney\}@syr.edu.}
}
%\author{Geethu~Joseph, M.~Cenk~Gursoy, and Pramod~K.~Varshney~\IEEEmembership{Life~Fellow,~IEEE}% <-this % stops a space
%\thanks{The authors are with the Department
%of Electrical Engineering and Computer Science, Syracuse University, Syracuse, NY 13244, USA e-mails: \{gjoseph,mcgursoy,varshney\}@syr.edu.}
%}

\maketitle

\begin{abstract}
We address the problem of monitoring a set of binary stochastic processes and generating an alert when the number of anomalies among them exceeds a threshold. For this, the decision-maker selects and probes a subset of the processes to obtain noisy estimates of their states (normal or anomalous). Based on the received observations, the decision-maker first determines whether to declare that the number of  anomalies has exceeded the threshold or to continue taking observations. When the decision is to continue, it then decides whether to collect observations at the next time instant or defer it to a later time. If it chooses to collect observations, it further determines the subset of processes to be probed. To devise this three-step sequential decision-making process, we use a Bayesian formulation wherein we learn the posterior probability on the states of the processes. Using the posterior probability, we construct a Markov decision process and solve it using deep actor-critic reinforcement learning. Via numerical experiments, we demonstrate the superior performance of our algorithm compared to the traditional model-based algorithms.
\end{abstract}

\begin{IEEEkeywords}
Active hypothesis testing, anomaly detection, change-point detection,
deep reinforcement learning, actor-critic algorithm, dynamic decision-making, sequential
sensing.
\end{IEEEkeywords}

\IEEEpeerreviewmaketitle

\section{Introduction}
Real-time monitoring (or diagnosis) applied to remote health monitoring, wireless sensor diagnosis, structural health monitoring, etc., is becoming ubiquitous today~\cite{lee2009wireless,chung2006remote,joseph2019anomaly}. Most such systems consist of several sensors of different modalities that gather data from the environment. Using this data, the diagnosis mechanism periodically checks the system health and alerts the system manager if the number of anomalies or outliers among the monitored quantities exceeds the maximum acceptable level. The anomalies can occur due to various reasons, and therefore, we use a stochastic framework to model the states of the processes and their evolution over time. Specifically, we assume that the state takes value 0 when the process is normal and 1 when the process is anomalous. Mathematically, this problem is equivalent to monitoring a set of stochastic binary processes and detecting when the number of processes with the state value 1 exceeds a predefined threshold. To monitor the processes, the decision-maker probes them to obtain an estimate of their states. This estimate is noisy, and it can differ from the true state with a certain probability. Therefore, the decision-maker needs to collect sufficient number of observations before it decides to issue an alert, i.e., the number of anomalies exceeds the threshold. However, probing all the processes all the time incurs a high sensing or communication cost, and so the decision-maker probes a small subset of processes at selected time instants.  Overall, the decision-maker has three decisions to make: detecting if the number of anomalies has exceeded the threshold, determining whether or not to probe at the next time instant, and choosing the subset to be probed.  Our goal here is to devise such a decision-maker which incurs minimum delay and cost and ensures the required level of accuracy. 

\subsection{Related Literature}
The two well-studied research topics related to our problem are change-point detection and active hypothesis testing. We briefly review them below.

\subsubsection{Change point detection} 
Change-point detection is the problem of observing a time series and finding abrupt changes in the data when a property of the time series changes. In our setting, the time series corresponds to the states of the monitored processes, and the changing property of the data is the number of anomalies. Numerous change-point detection algorithms, including various supervised and unsupervised methods, have been studied in the literature~\cite{naghshvar2010information,cleland2014evaluation,han2012comprehensive,
aminikhanghahi2017survey,liu2013change,agudelo2020bayesian,erdemir2021active}. The key difference between the classical change point detection problem and our setting is the controlled sensing part. Therefore, our sequential decision making is more complex because at every time instant, the decision maker not only determines whether or not the change has occurred, but also decides on which processes to be probed next.

\subsubsection{Active hypothesis testing} 
The sequential process selection problem is addressed by using the active hypothesis testing framework with multiple hypotheses. Specifically, in the context of anomaly detection, a hypothesis corresponds to a possible state of the processes~\cite{zhong2019deep, joseph2020anomaly}. There is a plethora of studies related to active hypothesis testing that explore the model-based approaches in the literature~\cite{chernoff1959sequential,bessler1960theory,nitinawarat2013controlled,naghshvar2013active,
huang2018active}. Recently, the active hypothesis testing framework is combined with deep learning algorithms to design data-driven anomaly detection algorithms~\cite{kartik2018policy,zhong2019deep,joseph2020anomaly,joseph2020anomaly2}. However, all these algorithms assume a static setting where the states of the processes remain unchanged throughout the decision-making process. In this paper, we consider a more challenging and practically relevant setting with Markovian state transitions over time. We set a different objective of determining whether or not the number of anomalies has exceeded the threshold.

\subsection{Our Contributions}
The original contributions of our paper are two-fold: \emph{time-varying hypothesis model and detection strategy}. To elaborate, we study a time-varying stochastic system wherein the system states evolve with time and we intend to detect a change point. For this, we note that the anomaly detection problem involves a trade-off between false alarms, detection delay, and sensing cost. Our Bayesian formulation relies on a stopping rule that continues to collect observations until the belief on the anomalous event (i.e., the event that the number of anomalies exceeds the acceptable level) is greater than a threshold. This approach controls the stopping time. Further, we formulate the dynamic selection of processes to be probed as a Markov decision process (MDP) whose reward function has two weighted terms. The first term controls the detection delay while the second term controls the sensing cost, and their weights balance the trade-off between the delay and cost. This novel reward formulation balances the trade-off between the performance metrics via the two user-defined parameters, namely the threshold on the belief and the regularizer (or weight). The MDP-based problem is then solved using the deep actor-critic reinforcement learning (RL) framework. The numerical results establish that our algorithm outperforms model-based decision making and offers higher flexibility in handling the trade-off between the performance metrics.

\section{Anomaly Detection and Tracking Problem}
We consider a set of $N$ stochastic processes where each process can be either normal or anomalous at each time instant. The state of a process can take two values: $0$ when the process is normal and $1$ when the process is anomalous. The states of all the processes at discrete time $t>0$ are represented using a vector $\vecs[t]\in\{0,1\}^N$, i.e., the $k\nth$ entry $\vecs_k[t]$ of $\vecs[t]$ corresponds to the state of the $k\nth$ process.  

The goal of the decision-maker is to monitor the processes until it detects the event $\calE[t]$ that the number of anomalies exceeds a predefined threshold $\uN\in\{1,2,\ldots,N\}$ at time $t$, i.e., $\sum_{k=1}^N\vecs_k[t]\geq\uN$.  To detect $\calE[t]$, the decision-maker is allowed to probe a subset of the $N$ processes at every time instant. It makes a three-step decision at any given~$t$:
\begin{enumerate}[leftmargin=0.3cm]
\item  Decide whether to declare the event $\calE[t]$ or to continue monitoring the processes.
\item If monitoring, decide whether to collect observations at time $t$ or at a later time $\tau>t$.
\item If collecting observations at time $t$, decide which subset $\calA[t]$ of the set of $N$ processes $\{1,2,\ldots,N\}$ to probe.
\end{enumerate}
Therefore, the decision-making algorithm has two parts: \emph{stopping rule} (mechanism to make the decision in Step 1), and \emph{selection policy} (mechanism to make the decisions in Steps 2 and 3). The other components of the system model and our design objectives are described below.
\subsection{State Transition}\label{sec:trans}
State transitions\footnote{The vector $\vecs[t]$ can take $m$ possible values which can be represented using $N-$bit representations of integers from 0 to $2^N-1$. For example, if $N=2$, then $\vecs[t]$ take values $\begin{bmatrix}
0&0
\end{bmatrix}\tran$, $\begin{bmatrix}
0&1
\end{bmatrix}\tran$, $\begin{bmatrix}
1&0
\end{bmatrix}\tran$, and $\begin{bmatrix}
1&1
\end{bmatrix}\tran$, which are the $2-$bit representations of $0,1,2,$ and $3$, respectively.}  over time are modeled using a \change{hidden} Markov chain defined by the \emph{state transition matrix} $\matP\in[0,1]^{m\times m}$. Here, we define $m\change{\triangleq}2^N$. The $(i,j)\nth$ entry $\matP_{ij}$ of the row stochastic matrix $\matP$ is given by
\begin{equation}\label{eq:MarkovModel}
\matP_{ij} = \bbP\lc \vecs[t]=\bin{}{j}\middle|\vecs[t-1]=\bin{}{i} \rc,
\end{equation}
where $\bin{}{i}\in\{0,1\}^N$ is the $N-$bit representation of integer $i-1$. 
%We note that $\sum_{j=1}^{m}\matP_{ij}=1$ for all values of $i$. 
Further, we also assume that if a process becomes anomalous, it does not go back to the normal state, i.e., 
\begin{equation}\label{eq:absorbing}
 \bbP\lc \vecs_k[\tau]=0\middle|\vecs_k[t]=1 \rc = 0,
\end{equation}
for any $k$ and $\tau>t$. Therefore, if the event $\calE[t]$ occurs at time $t$, then the event $\calE[\tau]$ continues to occur at all time $\tau>t$. 

\subsection{Observation Model}
When the decision-maker probes a subset $\calA[t]$ of processes, it obtains an estimate of their states, denoted by $\vecy_{\calA[t]}[t]\in\{0,1\}^{\lv\calA[t]\rv}$. These estimates or the entries of the observation vectors have a fixed probability $p$ of being erroneous, and the probability $p$ is called the flipping probability. Let $\vecy_k[t]\in\{0,1\}$ be the observation corresponding to the $k\nth$ process at time $t$. Then, we have%\footnote{While we consider a particular mapping in \eqref{eq:obs_model}, the analysis in the remainder of the paper follows similarly for any other random transformation between $s_k(t)$ and $y_k(t)$.  The only change is that the computation of the posterior probabilities would take into account the corresponding likelihood function of the observations.}
\begin{equation}\label{eq:obs_model}
\vecy_k[t] = \begin{cases}
\vecs_k[t] & \text{with probability } 1-p\\
1-\vecs_k[t] & \text{with probability } p.
\end{cases}
\end{equation}
Also, conditioned on the state, the observations  across different time instants are jointly (conditionally) independent, and thus, 
\begin{equation}\label{eq:indepen}
\bbP\lc \lc \vecy[\tau]\rc_{\tau=1}^t\middle|\lc \vecs[\tau]\rc_{\tau=1}^t\rc=\prod_{\tau=1}^t\prod_{k=1}^N\bbP\lc  \vecy_k[\tau]\middle| \vecs_k[\tau]\rc,
\end{equation}
for any $t>0$ with $\vecy[t]\!=\!\begin{bmatrix}
\vecy_1[t] & \vecy_2[t] \ldots \vecy_N[t]
\end{bmatrix}\tran\in\{0,1\}^N$.
\subsection{Design Objectives}\label{sec:goals}
%The objectives of this design depends on the following quantities:
%\begin{itemize}
%\item \emph{Change-point:} The smallest time instant at which the event $\calE[t]$ occurs is called the change-point $\tchange$. 
%\item \emph{Stopping time:} The time instant at which the decision-maker ends the observation acquisition phase and declares that the event $\calE[t]$ has occurred is called the stopping time $\tstop$. 
%\item \emph{False alarm:} If the stopping time is less than the change-point, it is called a false alarm.
%\item \emph{Detection delay:} If the stopping time is greater than the change-point, the difference between the two times $\tstop-\tchange$ is called the detection delay.
%\item \emph{Sensing cost:} The sensing cost is the total number of observations collected by the decision-maker before it makes a decision and it is given by $\sum_{t=1}^{\tstop}\lv\calA[t]\rv$.
%\end{itemize}

\change{We recall that our goal is to design a stopping rule and selection policy for detecting the event $\calE[t]$. Our algorithm assumes the knowledge of the system parameters $\matP$ and $p$ (that can be easily learned from the training data). The algorithm design} depends on three performance metrics: false alarm rate, detection delay $\tstop-\tchange$, and sensing cost $\sum_{t=1}^{\tstop}\lv\calA[t]\rv$. Here, $\tstop$ and $\tchange$ denote the stopping time (time when the observation acquisition phase ends) and change-point (time at which the event $\calE[t]$ starts occurring), respectively. 
Clearly, there is a trade-off between the three metrics. For example, if the decision-maker waits longer to collect more observations and reduce the false alarm rate, the detection delay and the sensing cost increase. In the following, we analyze the trade-offs between the metrics and derive a stopping rule and a selection policy that balance them.

\section{Stopping Rule}\label{sec:control}
This section derives the stopping rule of our algorithm. Our stopping rule relies on a metric called \emph{belief} (on the event $\calE[t]$) denoted by $\bbP_{\calE}[t]\in[0,1]$. The belief refers to the posterior probability of the event $\calE[t]$ conditioned on the observations available  to the decision-maker:
\begin{equation}\label{eq:belief}
\bbP_{\calE}[t] = \bbP\lc \sum_{k=1}^N\vecs_k[t]\geq\uN\middle| \lc \vecy_{\calA[\tau]}[\tau]\rc_{\tau=1}^t\rc.
\end{equation}
The stopping rule of the decision-maker is as follows:
\begin{equation}\label{eq:stopping}
\tstop = \min\lc t>0 : \bbP_{\calE}[t]>\upi\rc,
\end{equation}
where $\upi\in(0,1)$ is a predefined upper threshold on the belief. Here, due to the assumption in \eqref{eq:absorbing}, the (unconditioned) probability of the event $\calE[t]$ is a non-decreasing function of $t$, and so, we expect $\tstop$ to be finite. Also, the false alarm rate decreases with $\bbP_{\calE}[\tstop]$, and hence, our stopping rule controls the false alarm rate via $\upi$.
 
To compute $\bbP_{\calE}[t] $, we use its definition to arrive at
\begin{equation}\label{eq:E_update}
\bbP_{\calE}[t] = \sum_{i:\sum_{a=1}^{N}\bin{a}{i}\geq \uN}\vecpi_i[t],
\end{equation}
where $\bin{a}{i}$ is the $a\nth$ entry of the binary vector $\bin{}{i}\in\{0,1\}^{N}$. Also, we define $\vecpi[t] \in[0,1]^{m}$ as the posterior probability vector on $\vecs[t]$, and its $i\nth$ entry $\vecpi_i[t]$ is given by
\begin{align}\label{eq:pdf}
\vecpi_i[t] &= \bbP\lc\vecs[t]=\bin{}{i} \middle| \lc \vecy_{\calA[\tau]}[\tau]\rc_{\tau=1}^t\rc\\
&= \frac{\bbP\lc \lc \vecy_{\calA[\tau]}[\tau]\rc_{\tau=1}^t, \vecs[t]=\bin{}{i} \rc}{\sum_{j=1}^{m}\bbP\lc \lc \vecy_{\calA[\tau]}[\tau]\rc_{\tau=1}^t, \vecs[t]=\bin{}{j} \rc}.\label{eq:pdf1}
\end{align}
Here, we set $\calA[t]=\emptyset$ for the time instants at which the algorithm chooses not to probe any process. We further simplify \eqref{eq:pdf1} as follows:
\begin{align}
\bbP\lc \lc \vecy_{\calA[\tau]}[\tau]\rc_{\tau=1}^t, \vecs[t]=\bin{}{i} \rc\notag\\
&\hspace{-4.75cm}= \sum_{j=1}^{m}\bbP\lc \lc \vecy_{\calA[\tau]}[\tau]\rc_{\tau=1}^t, \vecs[t]=\bin{}{i},\vecs[t-1]=\bin{}{j} \rc\notag\\
&\hspace{-4.75cm}= \sum_{j=1}^{m}\bigg[\bbP\lc  \vecy_{\calA[t]}[t]\middle| \vecs[t]=\bin{}{i}\rc \matP_{ji} \vecpi_{j}[t-1]\notag\\
&\hspace{-0.8cm}\times   \bbP\lc\lc \vecy_{\calA[\tau]}[\tau]\rc_{\tau=1}^{t-1}\rc\bigg],\label{eq:inter1}
\end{align}
which follows from \eqref{eq:MarkovModel}, \eqref{eq:indepen} and \eqref{eq:pdf}. Also, from \eqref{eq:obs_model} and \eqref{eq:indepen},
\begin{multline}\label{eq:inter2}
\veceta_i[t]\triangleq \bbP\lc  \vecy_{\calA[t]}[t]\middle| \vecs[t]=\bin{}{i}\rc\\
= \prod_{a\in\calA[t]}p^{ \mathbbm{1}\lc \vecy_{a}[t]\neq\bin{a}{i}\rc}
(1-p)^{ \mathbbm{1}\lc \vecy_{a}[t]= \bin{a}{i}\rc},
\end{multline}
where $\mathbbm{1}$ is the indicator function. Also, we have
\begin{equation}
\sum_{j=1}^{m} \matP_{ji} \vecpi_{j}[t-1]=\vecpi[t-1]\tran\matP_{i},
\end{equation} where $\matP_{i}\in[0,1]^{m}$ is the $i\nth$ column of $\matP$. Thus, combining \eqref{eq:pdf1}, \eqref{eq:inter1}, and \eqref{eq:inter2}, we derive the recursive relation:
\begin{equation}\label{eq:pi_update}
\vecpi_i[t] = \frac{\vecpi[t-1]\tran\matP_{i}\veceta_{i}[t]  }{\vecpi[t-1]\tran\matP\veceta[t] },
\end{equation}
where $\veceta[t] = \begin{bmatrix}
\veceta_1[t] & \veceta_2[t] & \ldots & \veceta_{m}[t]
\end{bmatrix}\tran\in[0,1]^{m}$.
\change{We note that \eqref{eq:pi_update} is the same as the forward part of the forward-backward algorithm for a hidden Markov model.}

Combining \eqref{eq:stopping}, \eqref{eq:E_update}, \eqref{eq:inter2} and \eqref{eq:pi_update}, we obtain the stopping rule that maintains control over the false alarm rate via $\upi$. We next address the trade-off between the detection delay and sensing cost and investigate the selection policy that balances this trade-off. 

\section{Selection Policy}\label{sec:policy}
The selection policy chooses (or decides on) a subset $\calA[t]$ from the $m$ subsets (including the empty set) of the set of $N$ processes $\{1,2,\ldots,N\}$. This policy design is a dynamic decision making problem, and the best choice of $\calA[t]$ depends critically on the future states. An efficient approach to solve such problems is to utilize RL algorithms, and here, we use the policy-gradient RL algorithm called the actor-critic algorithm.

The goal of the RL algorithms is to sequentially choose actions in a dynamic environment to maximize the notion of a cumulative reward. The dynamic environment and reward are defined using the framework of MDP. Therefore, we first map our policy design problem to an MDP as presented next. 
\subsection{Markov Decision Process}
The problem of process subset selection can be modeled as an MDP with an associated set of actions and a reward function. Depending on the state of the MDP, the decision-maker chooses an action to receive the corresponding reward. It is easy to see that the state and the action of the MDP associated with the policy design are the process state $\vecs[t]$ and the set of selected processes $\calA[t]$, respectively. Consequently, the MDP state transitions are defined by the state transition matrix $\matP$ in \eqref{eq:MarkovModel}. However, we note that the MDP state (same as the process state) is not available to the decision-maker. Therefore, it chooses the actions based on the posterior probability on the state, %i.e., $\vecpi[t]$. Hence,  the action chosen at time $t$ (same as the policy) is a function of the posterior probability on the state computed at the previous time instant, 
i.e., $\calA[t]$ is a function of $\vecpi[t-1]$.

We complete the MDP framework by constructing a suitable reward function. Recall from \Cref{sec:goals} that we seek a policy that minimizes the false alarm rate, detection delay, and sensing cost, and the stopping rule in \Cref{sec:control} ensures that the false alarm rate is controlled via the parameter $\upi$. To further balance the trade-off between the detection delay and sensing cost, we introduce a regularizer or weight parameter $\lambda>0$ and define the instantaneous reward of the MDP as
\begin{equation}\label{eq:imm_reward}
r[t] = L\lb \bbP_{\calE}[t]\rb - L\lb \bbP_{\calE}[t-1]\rb -\lambda\lv\calA[t]\rv,
\end{equation}
where $L(x)=\log \ls x/(1-x)\rs $ is the log-likelihood function. The term $L\lb \bbP_{\calE}[t]\rb - L\lb \bbP_{\calE}[t-1]\rb$ in $r[t]$ encourages the decision-maker to choose actions that maximize the difference between the log-likelihoods of the events $\calE[t]$ and $\calE[t-1]$. This term motivates fast belief building, and thus, it lowers the detection delay. The remaining term $-\lambda\lv\calA[t]\rv$ forces the decision-maker to probe a small number of processes, and hence, it reduces the sensing cost. Hence, the reward balances the trade-off between the delay and cost via  the regularizer $\lambda$. 

Our algorithm aims at policies that maximize the long-term discounted reward given by $\bar{R}[t] = \sum_{\tau=t}^{\tstop}\gamma^{\tau-t} r[\tau],$
%\begin{equation}\label{eq:longterm}
%\bar{R}[t] = \sum_{\tau=t}^{\tstop}\gamma^{\tau-t} r[\tau],
%\end{equation} 
where $\gamma\in(0,1)$ is the discount factor. In other words, a reward received $\tau$ time steps in the future is worth only $\gamma^\tau$ times what it would be worth if it were received immediately. Thus, this formulation forces the algorithm to arrive at a decision faster, and thus, it minimizes the detection delay. The design of the policy that maximizes the long-term average reward using the actor-critic RL algorithm is presented next.

\subsection{Deep Actor-Critic Algorithm}
The deep actor-critic algorithm comprises two neural networks called the actor and critic neural networks.
The actor learns a selection policy to be followed by the decision-maker, and the critic predicts the expected reward the decision-maker receives if the actor's policy is followed. The actor learns from the prediction and improves its policy. The critic also updates its prediction by minimizing the error between the prediction and the actual reward received. After a few rounds of interactions between the two networks, the algorithm converges to the desired selection policy. In the following, we describe the actor-critic algorithm in detail.

\subsubsection{Network Architectures}
The actor neural network represents the stochastic policy that the decision-maker follows. It takes $\vecpi[t-1]\in[0,1]^{m}$ as the input and outputs a probability vector $\mu\lb\vecpi[t-1];\vecalpha\rb\in[0,1]^{m}$. The $i\nth$ entry of $\mu\lb\vecpi[t-1];\vecalpha\rb$ denotes the probability of choosing the $i\nth$ subset as $\calA[t]$, and  $\vecalpha$ denotes the set of parameters of the actor neural network. Thus, the decision-maker chooses $\calA[t]\sim \mu\lb\vecpi[t-1];\vecalpha\rb$. On the other hand, the critic neural network learns the expected reward or value function of the MDP, defined as 
%\begin{equation*}
$V(\vecpi) = \expect{\calA[t]\sim\mu\lb\vecpi;\vecalpha\rb}{\bar{R}[t]\middle|\vecpi[t-1]=\vecpi}$.
%\end{equation*}
Therefore, the input to the critic neural network is $\vecpi[t]\in[0,1]^{m}$, and the output is $V\lb \vecpi[t];\vecbeta\rb\in\bbR$. Here, $\vecbeta$ is the set of parameters of the critic neural network. We next discuss how the two networks adjust their parameters $\vecalpha$ and $\vecbeta$ to improve both policy and prediction.
\subsubsection{Network Updates}
At every time instant, the decision-maker chooses the subset $\calA[t]\sim\mu\lb\vecpi[t-1];\vecalpha\rb$, and obtains the observation vector $\vecy_{\calA[t]}[t]$. It then computes the immediate reward $r[t]$ using \eqref{eq:imm_reward}. From Bellman's equation~\cite[Chapter 3]{sutton2018reinforcement}, the error (called as temporal difference error) in the value function prediction can be computed as 
\begin{equation}\label{eq:temporalerror}
\delta[t] = r[t]+\gamma V\lb \vecpi[t];\vecbeta\rb-V\lb \vecpi[t-1];\vecbeta^{-}\rb\in\bbR,
\end{equation}
where $\vecbeta^{-}$ is the set of the critic neural network parameters obtained at the previous time instant. The actor aims at policies that maximize the value function, and the update of $\vecalpha$ can be derived as~\cite[Chapter 13]{sutton2018reinforcement}
\begin{equation}\label{eq:policy_gradient}
\vecalpha = \vecalpha^-+\delta[t]\nabla_{\vecalpha}[\log\mu_{a[t]}(\vecpi[t-1];\vecalpha)],
\end{equation}
where $\vecalpha^{-}$ is the set of the actor neural network parameters obtained at the previous time instant. Also, $\nabla_{\vecalpha}$ is the gradient with respect to $\vecalpha$, and $\mu_{a[t]}(\vecpi[k-1];\vecalpha)$ denotes the entry of $\mu(\vecpi[k-1];\vecalpha)$ corresponding to $\calA[t]$. At the same time, the critic minimizes the error in prediction and therefore, it updates $\vecbeta$ by minimizing $\delta^2[t]$ which is a function of $\vecbeta$.

The two neural networks alternately update their parameters to arrive at the selection policy that maximizes the long term reward. We summarize the overall algorithm in \Cref{alg:ActorCritic}.

\begin{algorithm}[hptb]
\caption{\strut Actor-critic RL for anomaly detection}
\label{alg:ActorCritic}
\begin{algorithmic}[1]
\REQUIRE $\upi\in(0,1)$, $\uN\in\{1,2,\ldots,N\}$, $\lambda\geq 0$, $\gamma\in(0,1)$

\ENSURE $\vecalpha,\vecbeta$ with random weights and $\vecpi[0]$ with the prior learned from the training data

\FOR {Episode index $= 1,2,\ldots$}
\STATE Time index $t =1$
\REPEAT 
\STATE Choose a subset $\calA[t]\sim\mu(\vecpi[t-1],\vecalpha)$ 
\STATE Receive observation $\vecy_{\calA[t]}[t]$
\STATE Compute probability vector $\veceta[t]$ using \eqref{eq:inter2} 
\STATE Compute posterior probability vector $\vecpi[t]$ using \eqref{eq:pi_update} 
\STATE Compute belief in decision $\bbP_{\calE}[t]$ using \eqref{eq:E_update}
\STATE Compute instantaneous reward $r[t]$ using \eqref{eq:imm_reward}
\STATE Update the actor network parameters $\vecalpha$ using \eqref{eq:policy_gradient}
\STATE Update the critic network parameters $\vecbeta$ by minimizing $\delta^2[t]$ in \eqref{eq:temporalerror}
\STATE Increase time index $t=t+1$
\UNTIL {$\bbP_{\calE}[t]>\upi$}
\STATE Declare the event $\calE[t]$
\ENDFOR
\end{algorithmic}
\end{algorithm}

\begin{figure*}[hptb]
\begin{center}
\begin{subfigure}{4.45cm}
\includegraphics[width= 4.45cm]{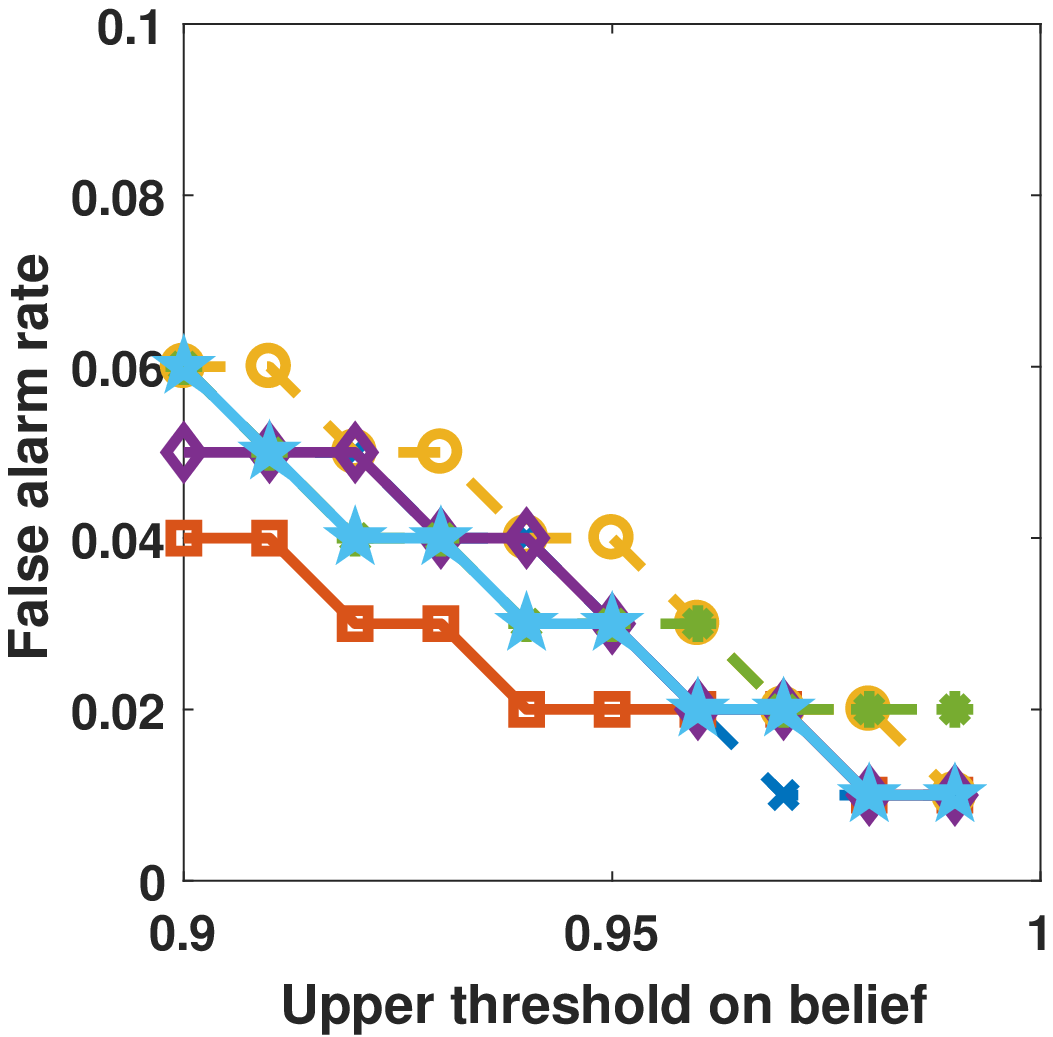}
\caption{}
\label{fig:FalseAlarm1}
\end{subfigure}
\begin{subfigure}{4.45cm}
\includegraphics[width= 4.45cm]{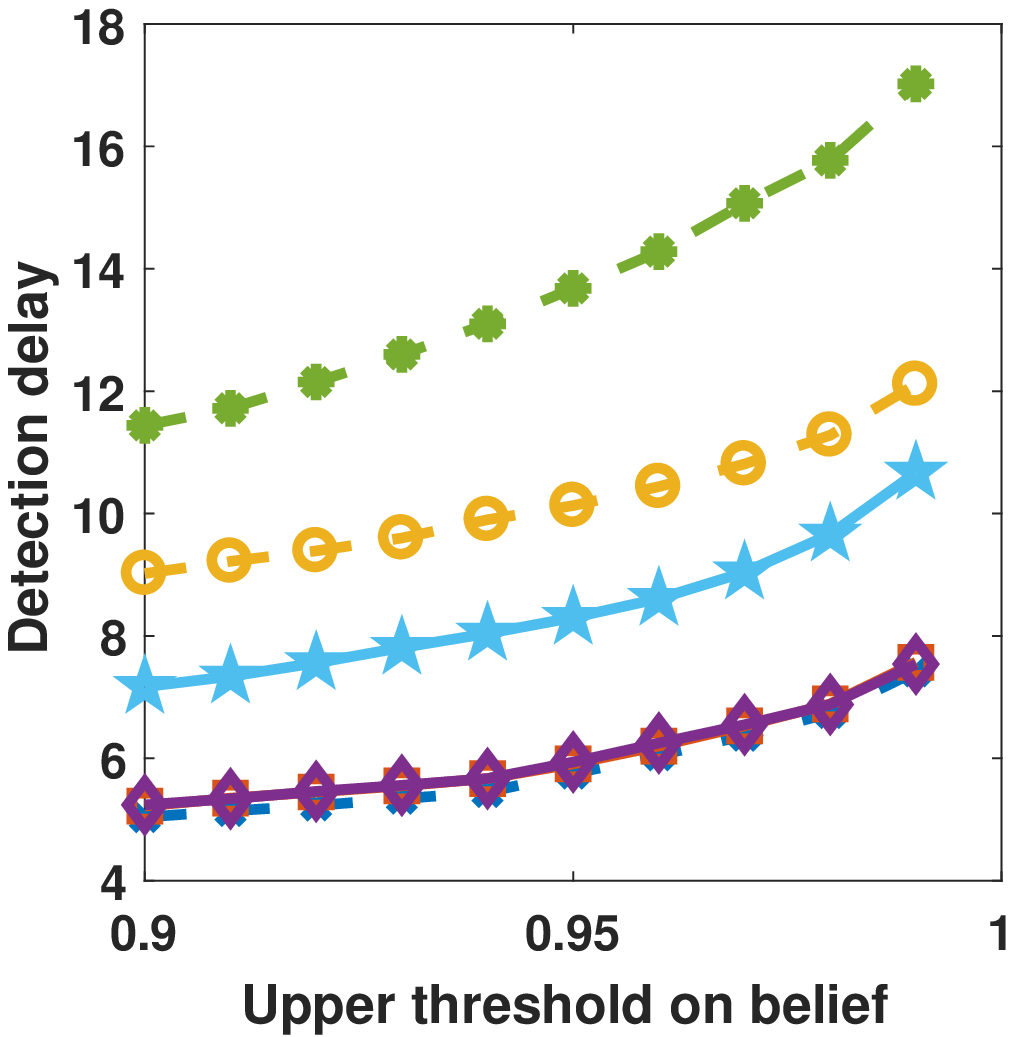}
\caption{}
\label{fig:Delay1}
\end{subfigure}
\begin{subfigure}{4.45cm}
\includegraphics[width= 4.45cm]{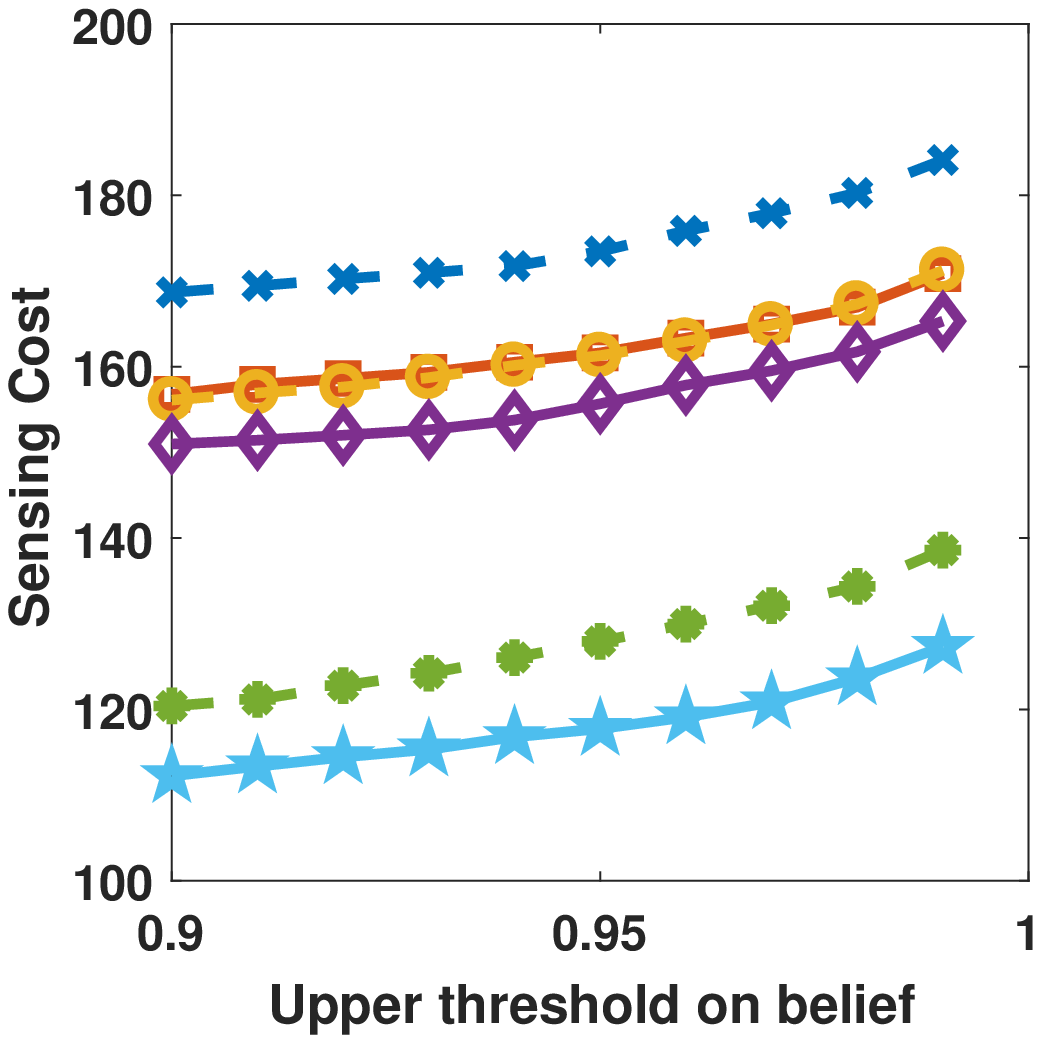}
\caption{}
\label{fig:Cost1}
\end{subfigure}
\begin{subfigure}{4.45cm}
\includegraphics[width= 4.45cm]{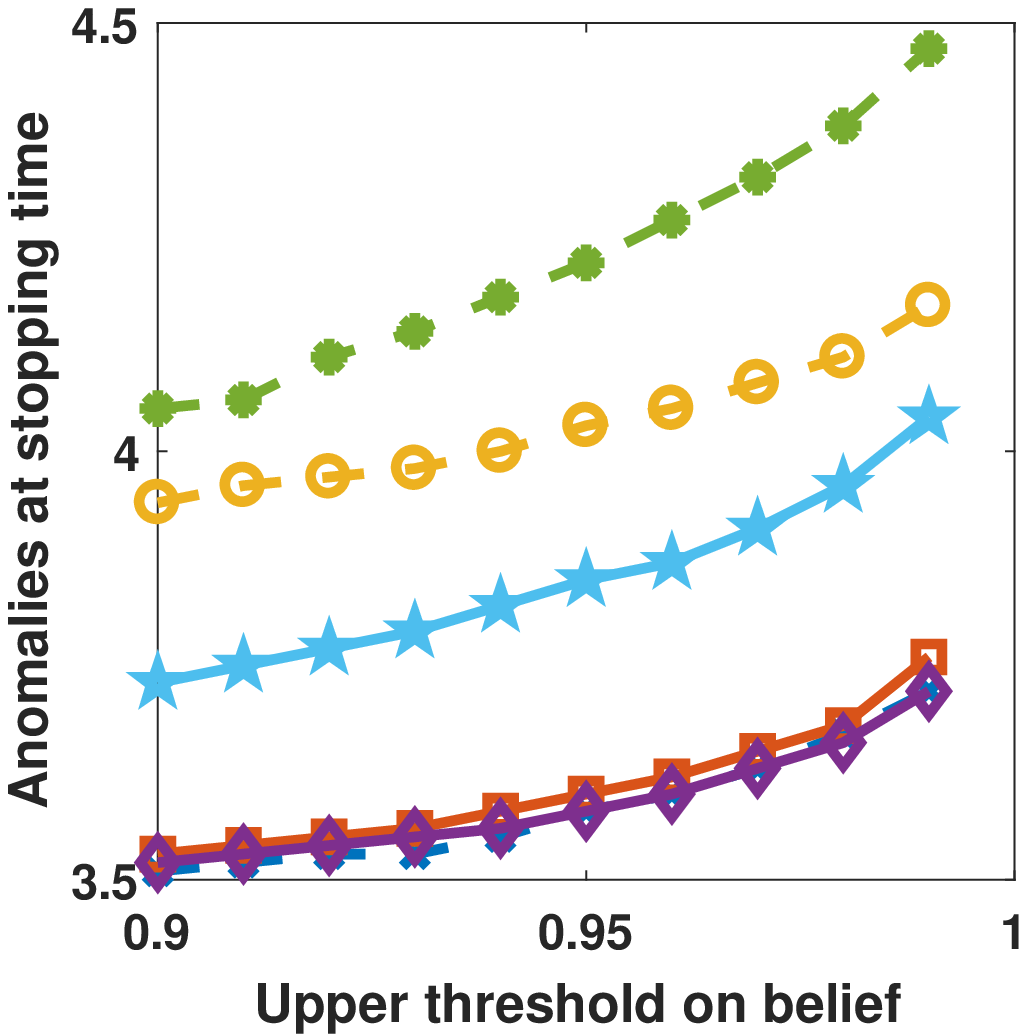}
\caption{}
\label{fig:Count1}
\end{subfigure}
%\begin{subfigure}{4.45cm}
%\includegraphics[width= 4.45cm]{EstAccuracy1}
%\end{subfigure}

\begin{subfigure}{\linewidth}
\centering
\includegraphics[height= 0.4cm]{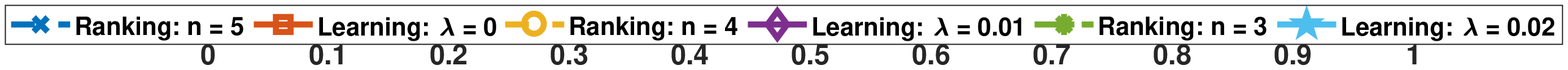}
\end{subfigure}

\end{center}

\caption{Performances of our deep actor-critic algorithm (labeled as \texttt{Learning}) and ranking-based algorithm (labeled as \texttt{Ranking}) as a function of $\upi$ for small values of $\lambda$ and large values of $n$ when $N=5$ and $\uN=3$.}
\label{fig:small}
\end{figure*}

\begin{figure*}[hptb]

\begin{center}
\begin{subfigure}{4.45cm}
\includegraphics[width= 4.45cm]{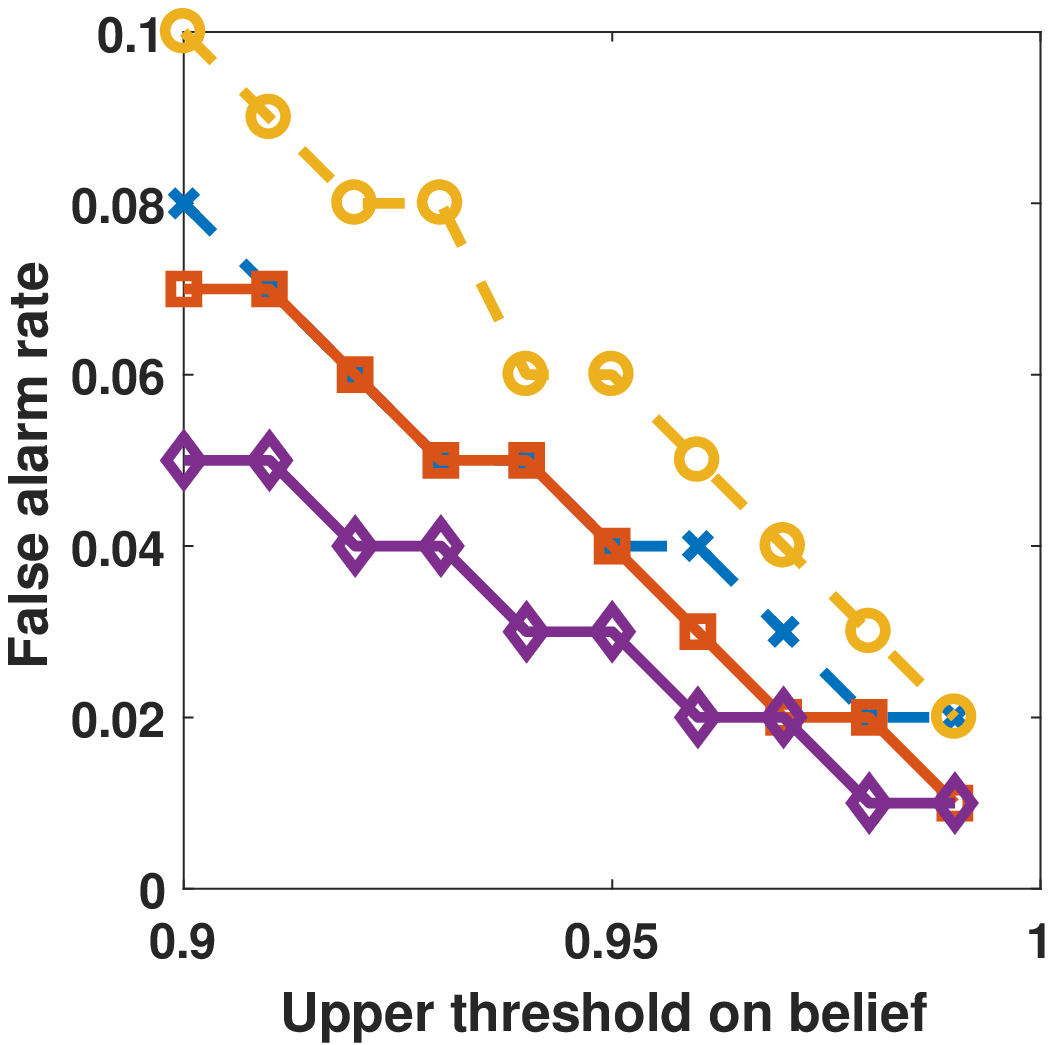}
\caption{}
\label{fig:FalseAlarm2}
\end{subfigure}
\begin{subfigure}{4.45cm}
\includegraphics[width= 4.45cm]{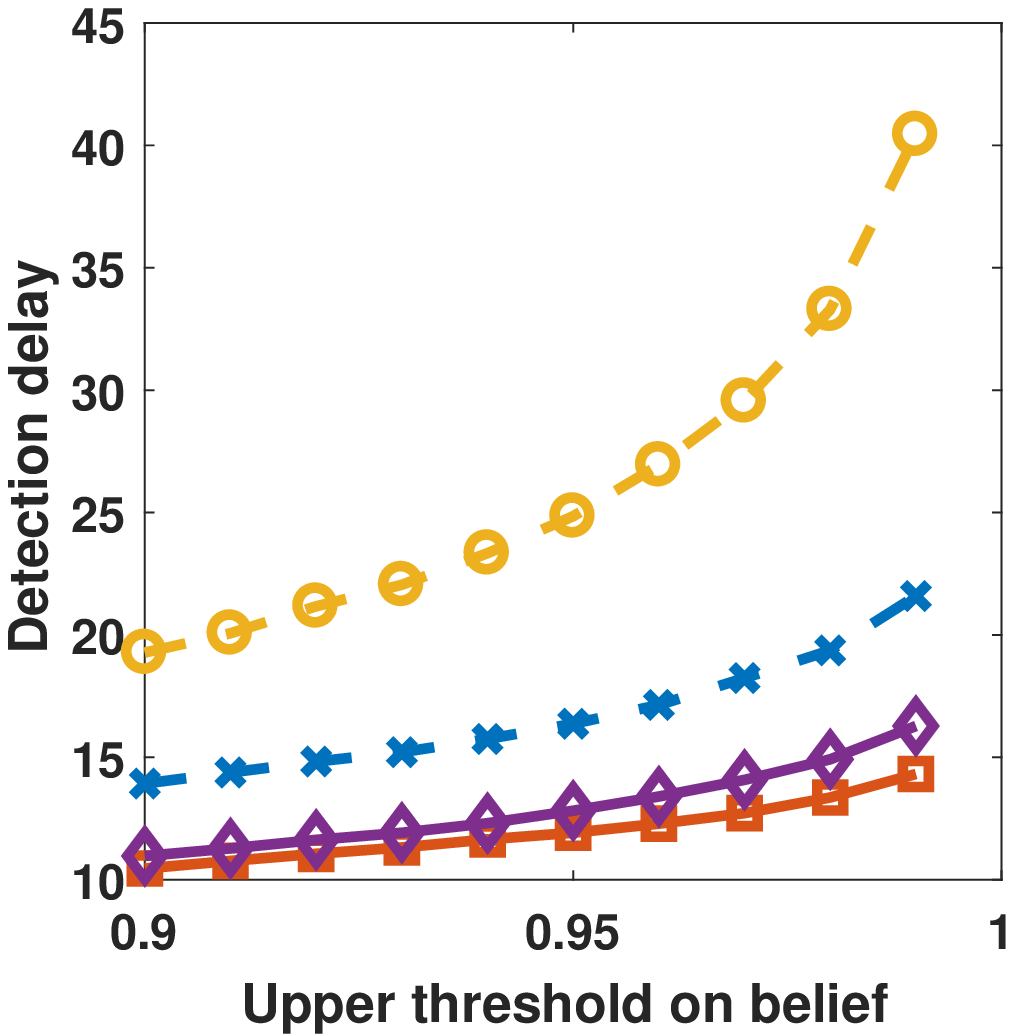}
\caption{}
\label{fig:Delay2}
\end{subfigure}
\begin{subfigure}{4.45cm}
\includegraphics[width= 4.45cm]{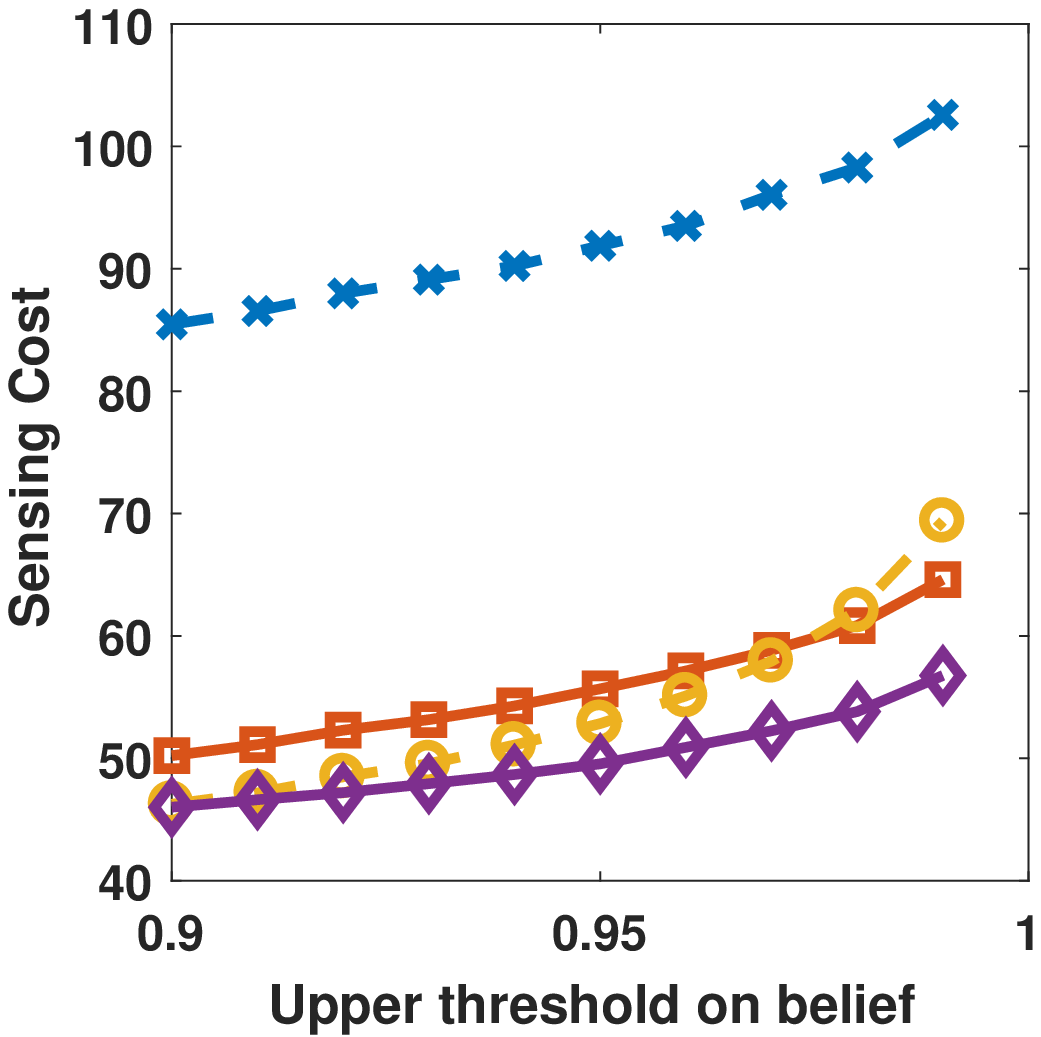}
\caption{}
\label{fig:Cost2}
\end{subfigure}
\begin{subfigure}{4.45cm}
\includegraphics[width= 4.45cm]{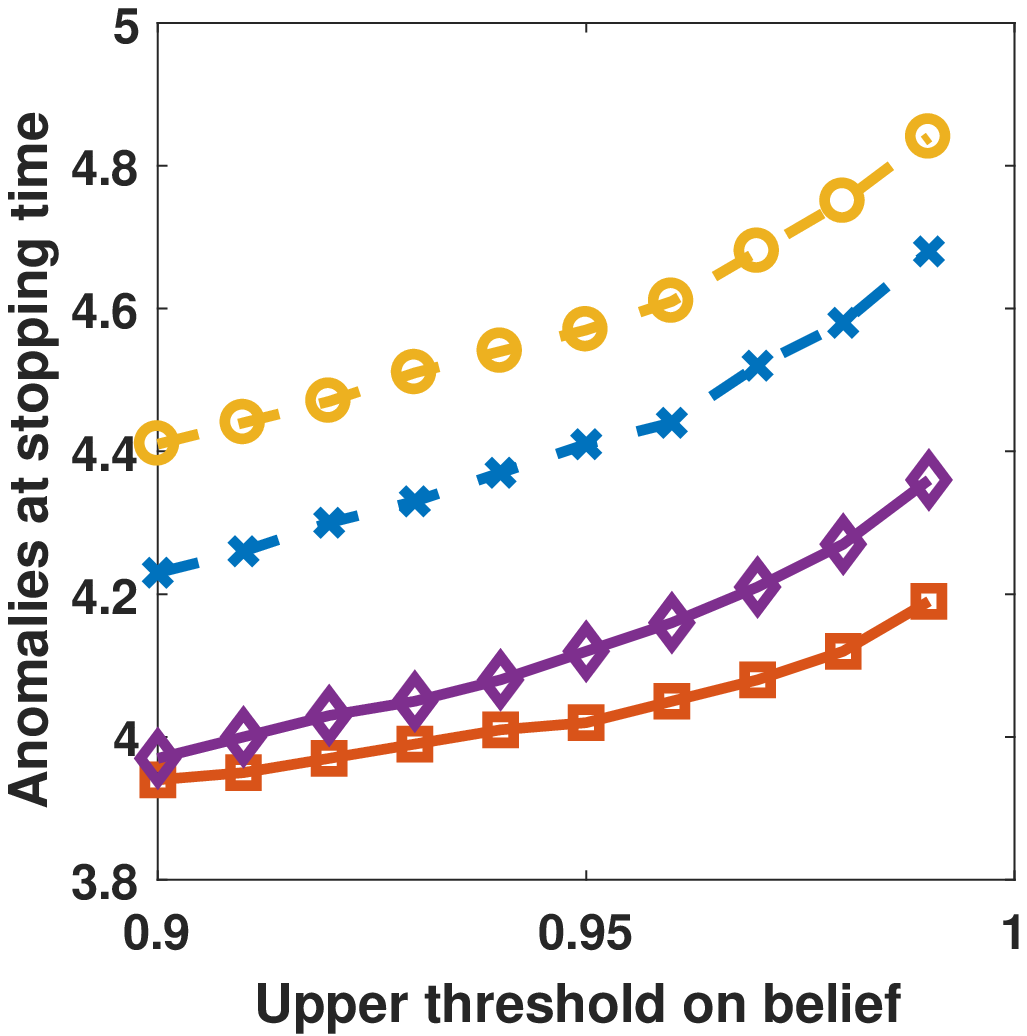}
\caption{}
\label{fig:Count2}
\end{subfigure}
%\begin{subfigure}{4.45cm}
%\includegraphics[width= 4.45cm]{EstAccuracy2}
%\end{subfigure}

\begin{subfigure}{\linewidth}
\centering
\includegraphics[height= 0.4cm]{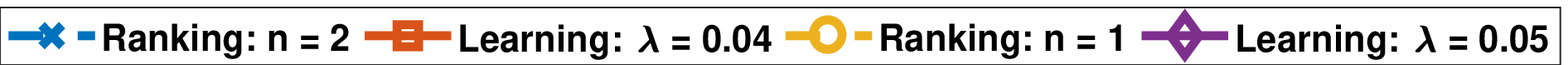}
\end{subfigure}
\end{center}

\caption{Performances of our deep actor-critic algorithm (labeled as \texttt{Learning}) and ranking-based algorithm (labeled as \texttt{Ranking}) as a function of $\upi$ for large values of $\lambda$ and small values of $n$ when $N=5$ and $\uN=3$.}
\label{fig:large}

\end{figure*}

\section{Simulation Results}\label{sec:simulations}

This section presents the numerical results illustrating the performance of our algorithm. We use four metrics to quantify the performance: false alarm rate, detection delay, sensing cost, and the number of anomalies among the processes at the stopping time i.e., $\sum_{k=1}^{N}\vecs_k[\tstop]$. We describe the simulation setup next.
\subsection{Simulation Setup}
We consider five processes, $N=5$ and set the threshold on anomalies as $\uN=3$.   Also, the flipping probability defined in \eqref{eq:obs_model}
is $p=0.2$. The other details are given below:

\subsubsection{State Transition} We assume that the process state is always initialized as the all-zeros state, i.e., $\vecs[0]=\zero\in\bbR^{N}$, and $\vecpi[0]=\begin{bmatrix}
1&0&0\ldots&0
\end{bmatrix}\tran\in\bbR^{m}$. Further, at every time instant, at most one process can move from the normal to the anomalous state while satisfying \eqref{eq:absorbing}. Also, we  set $q=\bbP\lc \vecs[t]\neq \vecs[t-1]\rc=0.1$ for all values of $t$. Thus, the row stochastic matrix $\matP\in\bbR^{m\times m}$ is given by%has the following structure:
%\begin{equation*}
%\matP_{ij}\!=\!\begin{cases}
%1-q & \text{ if } j=i, i\neq m\\
%1 & \text{ if } j=i= m\\
%\frac{q}{N-\sum_{a=1}^{m}\bin{a}{i}} &\text{ if } \begin{cases}
%\sum_{a=1}^{m}\bin{a}{j}-\bin{a}{i}=1\\
%\mathrm{dist}\lc \bin{}{j},\bin{}{i}\rc=1\\
%i\neq m
%\end{cases} \\
%0 & \text{ otherwise}.
%\end{cases}
%\end{equation*}
\begin{equation*}
\matP_{ij}\!=\!\begin{cases}
1-q & \text{ if } j=i, i\neq m\\
\frac{q}{N-\sum_{a=1}^{N}\bin{a}{i}} &\text{ if } j\in\calN\lc i\rc, i\neq m\\
1 & \text{ if } j=i= m\\
0&\text{ otherwise}.
\end{cases}
\end{equation*}
Here, $\calN\lc i\rc$ denotes the set of indices $j$ such that $\bin{}{j}$ is obtained by changing one of the zeros in $\bin{}{i}$ to one. %Also, we set the transition probability as $q=0.1$ in our experiments. 
 
\subsubsection{Implementation} We implement both actor and critic neural networks with three layers and use the ReLU activation function between their consecutive layers. The output of the actor neural network is normalized to unity to ensure that $\mu(\cdot)$ is a probability vector over the subsets of processes. The parameters of the neural networks are updated using the Adam Optimizer, and we set the learning rates of the actor and critic neural networks as $0.001$, and $0.05$, respectively. Also, we use a discount factor of $\gamma = 0.9$.

\begin{table}[b]
\caption{Number of processes probed per unit time as a function of the regularizer $\lambda$ when $N=5$ and $\uN=3$.}
\label{tab:lambda}

\begin{center}
        \begin{tabular}{|c|c|c|c|c|c|}
        \hline
\bf $\lambda$ & 0 &0.01&0.02&0.04&0.05\\
\hline
$\frac{1}{\tstop}\sum_{t=1}^{\tstop}\calA[t]$ &4.53 & 4.29 & 3.05 & 1.29 & 1.15\\
             \hline
        \end{tabular}
        
\end{center}

\end{table}

\subsubsection{Competing Algorithm} 
We compare our algorithm with a model-based algorithm that we refer to as the \emph{ranking-based algorithm}. This algorithm takes parameter $n\in\{1,2,\ldots,N\}$ which denotes the number of processes to be probed at each time instant, i.e., $\lv\calA[t]\rv=n$ for all values of $t$. At any given time $t$, the ranking-based algorithm chooses the top $n$ processes ranked in the order of their marginal probabilities $\lc \vecsigma_k[t]\rc_{k=1}^N $. Here, $\vecsigma_k[t]$ is the probability of the $k\nth$ process being anomalous at time $t$ when conditioned on the observations, and it is given by
\begin{equation*}
\sigma_k[t] = \bbP\lc\vecs_k[t]=1 \middle| \lc \vecy_{\calA[\tau]}[\tau]\rc_{\tau=1}^t\rc = \sum_{i:\bin{k}{i}=1}\vecpi_i[t].
\end{equation*}
%where $\vecpi\in[0,1]^{m}$ is defined in \eqref{eq:pdf}.

\subsection{Discussion of Results}
The insights from the simulation results presented in \Cref{fig:small,fig:large} and \Cref{tab:lambda} are as follows:

\subsubsection{Role of regularizer $\lambda$} From \Cref{fig:Delay1,fig:Cost1,fig:Delay2,fig:Cost2}, we see that as $\lambda$ increases, the detection delay of our RL-based algorithm  increases whereas its sensing cost decreases. This is intuitive from the immediate reward  in \eqref{eq:imm_reward} because $\lambda$ controls the weight of change in belief (the first two terms) and the cost term (third term) in $r[t]$. As a result, the policy becomes more cost-effective in terms of sensing for larger $\lambda$ and becomes more detection delay efficient for smaller~$\lambda$.

%\subsubsection{} 
%
%\subsubsection{Role of regularizer $\lambda$} From \Cref{fig:Delay1,fig:Cost1,fig:Delay2,fig:Cost2}, we see that as $\lambda$ increases, the detection delay of our RL-based algorithm  increases whereas its sensing cost decreases. This is intuitive from the immediate reward  in \eqref{eq:imm_reward} because as $\lambda$ increases, the cost term (third term) in $r[t]$ decreases. Thus, the algorithm selects fewer processes at every time instant (see \Cref{tab:lambda}). As a result, the policy becomes more cost-effective in terms of sensing. On the contrary, as $\lambda$ decreases, the first two terms in $r[t] $ gain more weight, and the policy becomes more detection delay efficient. Therefore, we conclude that $\lambda$ balances the trade-off between the detection delay and sensing cost. 

%\subsubsection{Comparing parameters $\lambda$ and $n$} 
\subsubsection{Comparing parameters $\lambda$ and $n$}  From \Cref{tab:lambda}, we notice that %when $\lambda=0$, the policy selects all the processes most of the time, and when $\lambda=0.05$, the policy selects only one process almost all the time. Therefore, 
$\lambda$ controls the number of processes probed per unit time whereas the ranking-based algorithm fixes this number via $n$. 
%This observation reveals that both algorithms balance the trade-off between the detection delay and sensing cost by the number of processes probed per unit time. However, 
We also observe from \Cref{fig:large,fig:small} that our algorithm (with a comparable number of processes probed per unit time) has better detection delay and sensing cost than the ranking-based algorithm for the same (or slightly better) false alarm rate. Further, $\lambda$ is positive real-valued whereas $n$ belongs to the finite set $\{1,2,\ldots,N\}$, and consequently, our algorithm is more flexible in handling the trade-off between detection delay and sensing cost.

\subsubsection{Role of threshold on belief $\upi$}  
 \Cref{fig:FalseAlarm1,fig:FalseAlarm2} show that %the false alarm rate decreases with $\upi$, and the false alarm rate of both algorithms under all settings are comparable for the same value of $\upi$. Therefore, 
$\upi$ controls the false alarm rate,  and the false alarm rate is less than $1-\upi$ for both algorithms under all settings. Also, from \Cref{fig:Delay1,fig:Cost1,fig:Delay2,fig:Cost2}, we infer that both detection delay and sensing cost increase with $\upi$ to meet the higher accuracy levels (decided by $\upi$). %, we need more observations and hence, higher detection delay. 
Further, from \Cref{fig:FalseAlarm1,fig:FalseAlarm2} we observe that for a large value of $\lambda$ and small value of $n$, the false alarm rate is slightly higher compared to the setting where $\lambda$ is small and $n$ is high. A reason for this trend is that the change in the belief in decision $\bbP_{\calE}[t]-\bbP_{\calE}[t-1]$ is larger when $\lambda$ is small due to the reward function formulation in \eqref{eq:imm_reward}. Therefore, the algorithm may be stopping at a higher value of $\bbP_{\calE}[t]$ compared to $\upi$ when $\lambda$ is small. 

\subsubsection{Number of anomalies at stopping time} 
\Cref{fig:Count1,fig:Count2} show that the number of anomalies at the stopping time given by $\sum_{k=1}^{N}\vecs_k[\tstop]$, has a similar trend as that of the detection delay. This is because as detection delay increases, the process state $\vecs[t]$ potentially changes, and more processes become anomalous due to \eqref{eq:absorbing}. In particular, \Cref{fig:Count2} shows that for $n=1$ and $\lambda=0.05$, the process state has reached the absorbing state of all-ones. This change in the state is expected as the probability of $\vecs[t]$ being different from $\vecs[t-1]$ has a relatively large value of $q=0.1$ (the number of anomalies at stopping time decreases with~$q$). 

\section{Conclusion}
We considered the problem of monitoring a set of binary processes and detecting when the number of anomalies among them  exceeds a threshold. The decision-maker was designed to aim at minimizing the false alarm rate, detection delay, and sensing cost. We balanced the trade-offs between these three metrics using a belief threshold-based  stopping rule and a weighted reward function-based MDP. Our solution was implemented using deep actor-critic RL that led to better performance and flexibility than the model-based algorithms. An interesting direction for future work is to extend our algorithm to detect when the anomalies exceed the threshold and to concurrently estimate the anomalous processes. 
%\ifCLASSOPTIONcaptionsoff
%  \newpage
%\fi

\bibliographystyle{IEEEtran}
\bibliography{AnomalyTracking}

\end{document}

%% file: My_notations.tex
\crefname{app}{Appendix}{Appendices}
\crefname{cor}{Corollary}{Corollary}
\crefname{prop}{Proposition}{Proposition}
\crefname{lemma}{Lemma}{Lemma}
\crefname{defn}{Definition}{Definition}
\crefname{conj}{Conjecture}{Conjecture}
\crefname{exam}{Example}{Example}
\crefname{supp}{Supplemental Section}{Supplemental Section}
\crefname{figure}{Fig.}{Figs.}
\crefname{section}{Sec.}{Secs.}

\newcommand{\bs}{\boldsymbol}
\newcommand{\bb}{\mathbb}
\newcommand{\mcal}{\mathcal}

\newcommand{\zero}{\bs{0}}

\newcommand{\lb}{\left(}
\newcommand{\rb}{\right)}
\newcommand{\ls}{\left[}
\newcommand{\rs}{\right]}
\newcommand{\lc}{\left\{}
\newcommand{\rc}{\right\}}

\newcommand{\lv}{\left\vert}
\newcommand{\rv}{\right\vert}

\newcommand{\LRV}[1]{{\left\vert\kern-0.25ex\left\vert\kern-0.25ex\left\vert #1 \right\vert\kern-0.25ex\right\vert\kern-0.25ex\right\vert}}

\newcommand{\expect}[2]{\bb{E}_{#1}\lc#2\rc}

\newcommand{\nth}{^\mathsf{th}}
\newcommand{\tran}{^{\mathsf{T}}}

\newcommand{\matP}{\bs{P}}

\newcommand{\bbP}{\bb{P}}

\newcommand{\bbR}{\bb{R}}

\newcommand{\calA}{\mcal{A}}
\newcommand{\calB}{\mcal{B}}

\newcommand{\calE}{\mcal{E}}

\newcommand{\calN}{\mcal{N}}

\newcommand{\vecs}{\bs{s}}

\newcommand{\vecy}{\bs{y}}

\newcommand{\vecalpha}{\bs{\alpha}}
\newcommand{\vecbeta}{\bs{\beta}}
\newcommand{\vecpi}{\bs{\pi}}

\newcommand{\veceta}{\bs{\eta}}

\newcommand{\vecsigma}{\bs{\sigma}}